# Realtime CNN-based Keypoint Detector with Sobel Filter and CNN-based Descriptor Trained with Keypoint Candidates

Xun Yuan, Ke Hu, and Song Chen

*Abstract—* The local feature detector and descriptor are essential in many computer vision tasks, such as SLAM and 3D reconstruction. In this paper, we introduce two separate CNNs, lightweight SobelNet and DesNet, to detect key points and to compute dense local descriptors. The detector and the descriptor work in parallel. Sobel filter provides the edge structure of the input images as the input of CNN. The locations of key points will be obtained after exerting the non-maximum suppression (NMS) process on the output map of the CNN. We design Gaussian loss for the training process of SobelNet to detect corner points as keypoints. At the same time, the input of DesNet is the original grayscale image, and circle loss is used to train DesNet. Besides, output maps of SobelNet are needed while training DesNet. We have evaluated our method on several benchmarks including HPatches benchmark, ETH benchmark, and FM-Bench. SobelNet achieves better or comparable performance with less computation compared with SOTA methods in recent years. The inference time of an image of 640×480 is 7.59ms and 1.09ms for SobelNet and DesNet respectively on RTX 2070 SUPER.

## I. INTRODUCTION

Nowadays, convolutional neural networks have contributed significant improvements in many computer vision tasks, such as object detection, semantic segmentation, and stereo matching. In the past few years, there have been lots of works that exploit CNN to detect keypoints, such as [1, 2, 4]. However, compared with traditional methods like SIFT, those methods don't show any significant improvement in terms of repeating errors, while CNNs always require much more computation. In contrast, CNNs show great performance in terms of descriptor matching such as HardNet [3]. Some modern tasks, such as SLAM [25] and augmented reality (AR) [26], attach great importance to real-time performance. As a result, dense descriptors that could be obtained through a one-time computation are preferred and patch descriptors like HardNet become less attractive owing to the time-consuming process which transforms the whole image to patch inputs of a certain size.

Handcraft filters and image pyramid are usually used in traditional keypoint detection methods, such as the Difference of Gaussians. The latter one is used to detect keypoints of different scales. In terms of computing descriptors, traditional methods often compare the keypoints with their neighbor points to determine the directions and values of their corresponding descriptors. In recent years, many researchers

Xun Yuan is with University of Science and Technology of China, Hefei, China (e-mail: yx111@mail.ustc.edu.cn).
Ke Hu is with University of Science and Technology of China, Hefei, China (e-mail: kehu@mail.ustc.edu.cn).
Song Chen is with University of Science and Technology of China, Hefei, China (corresponding author to provide phone: 05516360 2675; e-mail: songch@ustc.edu.cn).

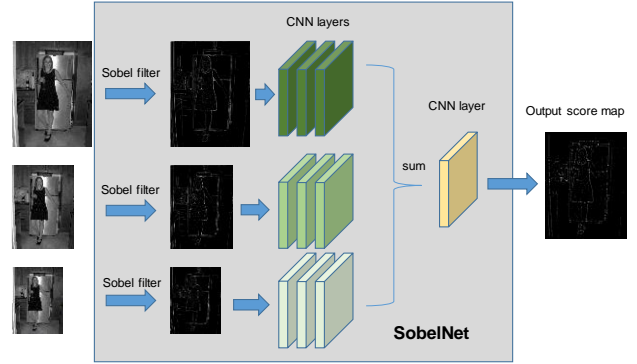

Fig. 1. The architecture of SobelNet contains Sobel filter and CNN layers. Different colors represent different CNN layers. The input image pyramid includes the original size image and two down scaled images whose scale factors are √2 and 2 respectively. And before summation, downscaled feature maps are upscaled to the original size.

attempt to utilize CNNs to detect keypoints and compute their descriptors. Most of them choose the original image as the input of their networks, such as [1, 4, 5]. However, our work is inspired by Key.Net [2], which uses the output feature maps of handcraft filters as the inputs of CNN. Handcraft filters could help to significantly reduce the computation burden of CNN because it eliminates most of the information in the images which is redundant as we just care about finding out the corner points. Similar to Key.Net, we also use multi-scale images as the inputs of SobelNet to detect the same points with small scale change. Different from Key.Net that uses several handcraft filters, our method only uses one handcraft filter, Sobel filter that is an easy and effective filter to detect the edge structures of an image and robust with the Gaussian noise in the image. We believe that CNN can be trained to find out corner points based on the edge structures and carefully designed loss function.

In recent years, training keypoint detector and descriptor in a joint manner, such as [1, 4, 5], is becoming a trend with the assumption that keypoints' locations have a close relationship with their corresponding descriptors. The network architectures of those works always contain a shared backbone network and two separate heads for keypoint detection and descriptor computation respectively. What's more, some works such as [8, 27] exploit the difference in descriptor map to train the detector, considering that points with similar descriptors should not be treated as keypoints. However, in our opinions, keypoint detection and descriptor computation are different tasks and should focus on separate issues. It is the obligation of the detector to detect the same keypoints with different viewpoints as many as possible. And corner points meet the above condition. In most cases, corner points will remain as corner points after a perspective transformation of an image. As a result, the purpose of SobelNet is to detect corner points. As for descriptor, the descriptors of the corresponding points between the images before and after a perspective transformation should be similar and the descriptors of the non-corresponding points should be distinctive.

There are three main contributions of this paper: 1) SobelNet exploits the Sobel filter to eliminate redundant information of images and reduce the computation requirement of CNN. 2) We propose a loss function, Gaussian loss, to train SobelNet to detect corner points as keypoints. 3) We use keypoint candidates to train DesNet to improve the performance of DesNet.

## II. RELATED WORKS

In terms of traditional keypoint detectors and descriptors, Scale-Invariant Feature Transform (SIFT) [7] is the most well-known method that has great performance in many computer vision tasks. SIFT detects initialed location of key points through multi-level Difference of Gaussian (DoG) process, then rejects keypoints with low contrast and eliminates edge responses to accurate keypoint localization. A keypoint's descriptor is created by computing the gradient magnitude and orientation at each image sample point in a region around the keypoint location. This is a time-consuming method. In real-time applications like [25], ORB [6] is much more popular, which improves the computing speed at the expense of detecting and matching accuracy.

As for machine learning methods, joint network architectures and joint learning methods attract much research attention, such as [1, 4, 5, 8]. SuperPoint [4] presents a self-supervised framework for training interest point detectors and descriptors which is pre-trained on Synthetic Shapes and then trained on real images by the Iterative Homographic Adaptation technique. That means it uses the current network to detect key points as pseudo-ground truth to train the network again. UnsuperPoint [5] is inspired by SuperPoint which produces a score map and a relative position coordinate map for keypoint detector rather than applying a "depth to space" method as SuperPoint which makes iterative training a necessary. However, UnsuperPoint meets trouble that it prefers to detect key points at the edge of the 8×8 region although it applies a forced method to make the position coordinates of key points distribute uniformly in the 8×8 region. R2D2 [8], whose detector produces a reliability map and a repeatability map, puts both repeatability and discriminant issues into consideration to detect keypoints. But as far as we are concerned, distinguishing the keypoints is the descriptors' job, and besides, a keypoint with a certain structure may be discriminative in one image but not in another image. As a result, it's hard to define discriminant in the keypoint detection process. Besides, ASLFeat [1], which contains DCN [28] in its network, performs great on several benchmarks. ASLFeat post-processes the keypoints with the SIFT-like edge elimination (with the threshold set to 10) and sub-pixel refinement, then the descriptors are bilinearly interpolated at the refined locations. As ASLFeat's architecture is already relatively big, it's hard to run it on a common hardware device in real-time.

On the contrary, [2, 9] consider that the keypoint detector and descriptor are separate tasks. LF-Net [9] detects key points and computer their descriptors in a sequence manner. Its detector network generates a scale-space score map along with dense orientation estimates, which are used to select the keypoint positions. Image patches around the chosen keypoints are cropped with a differentiable sampler (STN) and fed to the descriptor network, which generates a descriptor for each patch. However, patches generation is a time-consuming task and the receptive field is limited if only compute descriptors from a certain size image patches. Key.Net [2], which combines handcrafted and learned CNN filters within a shallow multi-scale architecture, concentrates only on the detector and utilizes HardNet [3] to compute descriptors. It performs better than previous joint learning methods on the HPatches benchmark.

## III. ARCHITECTURES AND LOSS FUNCTIONS

In this chapter, the architectures of SobelNet and DesNet will be introduced. And we also present our original Gaussian loss and the implementation of circle loss [10].

### A. Architecture of SobelNet

The overall framework of SobelNet is shown in Fig. 1. The input image pyramid includes the original size image and two downscaled images whose scale factors are √2 and 2 respectively. Each image is processed through three convolutional layers with eight channels remaining its dimension. After the last layer of the three layers, the downscaled feature maps are upscaled to the original image size. Subsequently, feature maps will be summed up to put the features of all scale images into consideration. Then, the summation is sent to the final convolutional layer to produce an output score map. We set the value of points in the score map whose values are less than a certain ratio of the maximum value of the score map to zero to discard the low quality keypoint candidates. In the end, a non-maximum suppression (NMS) is applied to detect final keypoints.

The network would be robust with scale changes owing to the image pyramid input. And besides, before downscaling the image, we process the image through a Gaussian filter which could contribute to reducing the Gaussian noise in the image.

The process of the Sobel filter can be written as:

$$filter\_x = \begin{bmatrix} -1 & 0 & 1 \\ -2 & 0 & 2 \\ -1 & 0 & 1 \end{bmatrix} \quad filter\_y = \begin{bmatrix} -1 & -2 & -1 \\ 0 & 0 & 0 \\ 1 & 2 & 1 \end{bmatrix}$$

$$Sobel\_out = \sqrt{filter\_x^2(img) + filter\_y^2(img)} \quad (1)$$

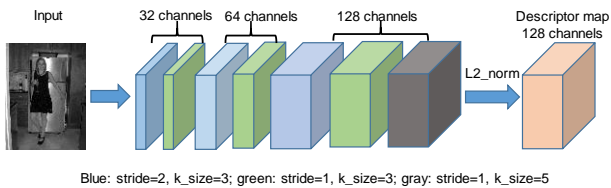

Blue: stride=2, k_size=3; green: stride=1, k_size=3; gray: stride=1, k_size=5

Fig. 2. Architecture of DesNet is shown in this figure.

where *img* and *Sobel_out* represent input grayscale image and output of Sobel filter respectively.

### B. Architecture of DesNet

In our opinion, a descriptor should have a large receptive field to contain as much information around the keypoints as possible. And if the receptive field is large enough, descriptors of different keypoints with a similar structure could be distinct enough to be distinguished, because their surrounding areas can't be all the same in most of the real-world images. DesNet has a very common CNN architecture which is shown in detail in Fig. 2.

### C. Gaussian Loss

Gaussian loss is designed to train SobelNet to detect corner points. Some details of Gaussian loss are shown in Fig. 3. When computing the Gaussian score of every point in the output score map (referred to Fig. 1), a Gaussian kernel, whose center value is normalized to 1, moves in X and Y directions with one stride. And the point which is being computed is always on the center of the Gaussian kernel so that there would be no edge effect, such as [4, 5], in our method. The gaussian score is computed as follow:

$$Gauss\_score_{center} = \frac{weight_{center} * p_{center}}{\sum_{i \in R} weight_i * p_i + \varepsilon} * p_{center} \quad (2)$$

where *R* represents the region in the image covered by the current Gaussian kernel and $weight_i$ and $p_i$ represent the value of the Gaussian kernel and the output score map in the corresponding position. Considering that a higher score always means a higher quality point, we multiply the fraction by the value of the center point.

Under ideal conditions, we assume that every point of the edge has an equal value as shown in Fig. 3c and Fig. 3d. The blue and green square in Fig. 3c and Fig. 3d denote the same Gaussian kernel computing on different positions of the edge, and the dots in the center of squares represent the computing points or center points in terms of output score map and Gaussian kernel respectively. In Fig. 3c it's clear that the Gaussian score of the blue dot is higher than that of the green dot since the part of edge locating in the green square is longer than that located in the blue square. However, it's hard to determine which one is longer as for the circumstance such as Fig. 3d. But Gaussian kernel has larger weights in its center area than those in its edge area and the edge is always more than 2 pixels wide as for output of Sobel filter. As a result, in the green square of Fig. 3d, there are more points in its central areas so that in most cases, the Gaussian score of the blue dot is higher than the green one. Overall, the Gaussian score is competent to distinguish edge points and corner points. In addition, the characteristic of the Gaussian kernel also helps to reduce the impact of other edges or noise points in its edge area.

Now we have Gaussian scores of each point in the output score map. Subsequently, the Gaussian score map will go through an NMS process to find out local maximum scores and then local maximum scores are set 1 if they are greater than zero, while the others are set 0 to obtain corner point map. Then we design a mask to filter out low-quality corner points as follows:

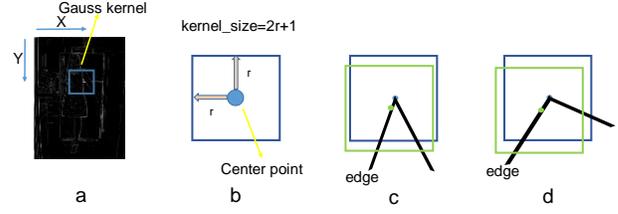

Fig. 3. Some details about Gaussian loss are shown in this figure. Figure (a) presents a Gaussian kernel sliding across the output score map with stride one. Figure (b) shows the structure of Gaussian kernel whose center point corresponds to the point to be computed in the output score map. Figure (c) and figure (d) present two circumstances when computing Gaussian loss at corner point and edge point.

$$mask = \begin{cases} 1 & where \quad sobel\_edge > \alpha * \max(sobel\_edge) \\ 0 & otherwise \end{cases} \quad (3)$$

so that the impact of noise in the input image and produced in the computing process could be eliminated. The new corner point map is obtained by multiplying the corner point map and mask:

$$corner\_point\_map = corner\_point\_map * mask \quad (4)$$

The Gauss loss for the point is calculated by the following equation:

$$Gauss\_loss\_point = 1 - \frac{weight_{center} * p_{center}}{\sum_{i \in R} weight_i * p_i} \quad (5)$$

where the right part of (5) is the same as the left part of (2). We only focus on the Gaussian loss of the points whose values in corresponding positions of corner point map are non-zero. In the end, the final Gaussian loss is the sum of Gaussian loss for points multiplied with normalizing weights as follows:

$$weight_i = \frac{Gauss\_score\_point_i}{\sum_{i \in R} Gauss\_score\_point_i + \varepsilon} \quad (6)$$

$$Gauss\_loss = \sum_{i \in R} Gauss\_loss\_point_i * weight_i \quad (7)$$

where *R* is a set of positions where the points' values are non-zero in the corner point map.

### D. Implementation of Circle Loss

During the training process of DesNet, we utilize circle loss [10] to train our network. We set the two hyparameters in circle loss *m* and $\gamma$ to 0.1 and 1. Please refer to circle loss [10] for more details.

## IV. TRAINING PROCESS

In this section, we will explain the way how we train SobelNet and DesNet in detail and introduce the datasets we use during the training process.

### A. Training and valuation datasets

We select 12044 images whose height and width are both longer than 480 pixels from the MS COCO [29] 2017 training dataset and 4000 images randomly selected from the TUM [18] dataset as the training dataset of SobelNet. And 500 images are randomly selected from the TUM dataset as the valuation dataset of SobelNet.

As for DesNet, we select 12044 images whose height and width are both longer than 480 pixels from MS COCO [29] 2017 training dataset as training dataset of DesNet and 471 images whose height and width are both longer than 480 pixels from MS COCO [29] 2017 valuation dataset as valuation dataset of DesNet.

### B. Training process of SobelNet

As few datasets are designed specifically for keypoint detection task, we need to create our dataset. After we have chosen images for training, we modify the contrast, brightness, and hue value in HSV space to original images to improve the network's robustness against illumination changes as Key.Net [2]. Then we apply homography, rotation, and scale transformations to the original images in sequence.

First of all, four corners of the images randomly move [-0.15, 0.15] percentage of the length of height and width of the images in both "Y" and "X" directions. Then we rotate the current images by [-90, 90] degrees randomly that contribute to the robustness of the Gaussian loss since the Gaussian kernel is symmetry by 90 degrees. At last, we rescale the current images to [0.85, 1.15] of their original size.

The original images, transformed images, and transformation matrices are sent to the training process. After being processed by SobelNet, we obtain two output maps, the original score map (OSP) and the transformed score map (TSP). Subsequently, we transform OSP and TSP by the corresponding transformation matrix and get OSP-T and TSP-T. Now, we compute the loss function to be optimized by the following equations:

$$GL\_OSP = \sum_{i \in R_1} GLP\_OSP_i * weight\_OSP_i \quad (8)$$

$$GL\_TSP = \sum_{i \in R_2} GLP\_TSP_i * weight\_TSP_i \quad (9)$$

where $GL$ and $GLP$ are short for Gaussian loss and Gaussian loss for point respectively, and $R_1$ and $R_2$ represent a set of positions where the point values are non-zero in TSP-T's corner point map and a set of positions where the point values are non-zero in OSP-T's corner point map respectively.

In the end, we apply the multi-scale Gauss Loss to train SobelNet to obtain robust performance for all kinds of corner points. We set "r" of the three Gauss kernels 4, 6, and 8 respectively. The final Gauss loss is computed as follows:

$$GL\_final = \sum_{r \in \{4,6,8\}} GL\_OSP_r + GL\_TSP_r \quad (10)$$

### C. Training process of DesNet

In this section, it turns to train our DesNet. Our purpose is computing descriptors to distinguish every keypoint. As a result, it would be more effective to just train the descriptors of the points (keypoint candidates) whose corresponding output score map points' values are greater than $\alpha$ (a setting threshold), than those of random points of the input image. The benefit could be seen in Table I. SobelNet* is a contrast network whose DesNet is trained with randomly selected points. The performance of the descriptors of SobelNet* shows a considerable drop.

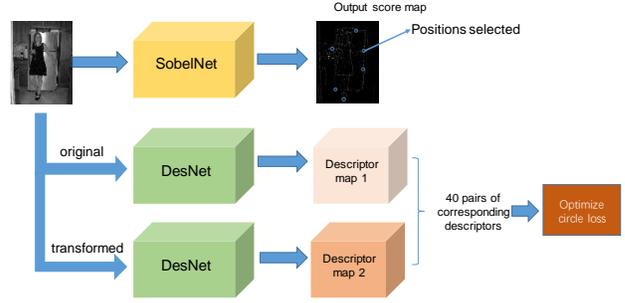

Fig. 4. The whole process of training DesNet is presented. Firstly, we obtain output score map and descriptor map 1 and 2 through SobelNet and DesNet. Then we select 40 positions according to output score map and choose the corresponding positions in both descriptor map 1 and 2. At last, circle loss of the 40 pairs corresponding descriptors is used to optimize our DesNet params.

And we randomly choose 40 descriptors of the positions in both descriptor map 1 and descriptor map 2 as shown in Fig. 4, that are apart from each other longer than 7 pixels and chosen from the potential area as introduced above. The transformation is the same as that in the training process of SobelNet, except that we don't apply any rotation here. Then circle loss is used to train the 40 pairs of descriptors, to make the descriptors of the corresponding positions more similar and the different positions more distinct. The whole process is shown in Fig. 4.

## V. EXPERIMENTS

In the following sections, we will introduce the metrics and the dataset used for evaluating our method. We evaluate our method on three benchmarks, namely HPatches benchmark [13], FM-Bench [12], and ETH benchmark [11]. We mainly compare our method against SIFT, a well-known traditional method, Key.Net, and ASLFeat which have the SOTA performance among machine learning methods in recent years.

### A. HPatches benchmark

HPatches benchmark, which is a large-scale dataset of image sequences annotated with homographies, is widely used as a benchmark for keypoint detection and descriptor computation tasks, such as [1, 2, 4, 5]. The benchmark contains 117 sequences. In 57 scenes, the main nuisance factors are photometric changes and the remaining 59 sequences show significant geometric deformations due to viewpoint change. In this benchmark, we mainly focus on three metrics: 1) Key point repeatability (%Rep), the ratio of possible matches, and the minimum number of key points in the shared view. 2) Matching score (%M.S.), the ratio of correct matches (error threshold @ 5 pixels), and the minimum number of key points in the shared view. 3) Mean matching accuracy (%MMA), the ratio of correct matches, and possible matches. "Kpts" in Table I means the average keypoint amount in the shared view of the corresponding images. In this section, we limit the maximum number of keypoints of all methods to 5K and set NMS size 15 for machine learning methods the same as [2]. The experiment results are shown in Table I. Our method performs better on %Rep while ASLFeat has better performance on %M.S and %MMA.

Table I   Comparison of Rep (%), M.S (%) and MMA (%) on HPatches benchmark

|  | Viewpoint | | | Illumination | | | Overall | | | Kpts |
| --- | --- | --- | --- | --- | --- | --- | --- | --- | --- | --- |
|  | Rep | M.S | MMA | Rep | M.S | MMA | Rep | M.S | MMA |  |
| SIFT (python) | 75.24 | 33.53 | 42.29 | 72.68 | 25.31 | 33.64 | 73.98 | 29.49 | 38.04 | 2467 |
| Key.Net [2] + HardNet [3] | 83.91 | 45.24 | 67.19 | 80.50 | 49.70 | 79.49 | 82.23 | 47.43 | 73.23 | 1430 |
| ASLFeat [1] | 79.94 | **52.03** | **76.76** | 77.49 | 49.61 | **81.59** | 78.74 | **50.84** | **79.13** | 1144 |
| SobelNet* | **85.15** | 45.12 | 65.09 | **82.44** | 48.51 | 76.03 | **83.82** | 46.79 | 70.46 | 1533 |
| SobelNet | **85.15** | 47.44 | 68.95 | **82.44** | 50.46 | 79.68 | **83.82** | 48.92 | 74.22 | 1533 |

Table II   Evaluation on FM-Bench

|  | TUM | | | | KITTI | | | |
| --- | --- | --- | --- | --- | --- | --- | --- | --- |
|  | %Recall | %Inlier(-m) | #Corrs(-m) | Kpts | %Recall | %Inlier(-m) | #Corrs(-m) | Kpts |
| SIFT (python) | 55.20 | 75.41 (65.63) | 95 (343) | 831 | **92.30** | 98.45 (88.39) | 337 (807) | 2764 |
| Key.Net[2]+HardNet[3] | 51.00 | 71.81 (62.97) | 217 (1208) | 2505 | 85.90 | 98.20 (94.28) | 290 (1447) | 4991 |
| ASLFeat [1] | 59.90 | 75.35 (69.46) | 102 (490) | 917 | 92.00 | 98.62 (96.82) | 370 (1212) | 3057 |
| SobelNet | **61.50** | 75.16 (65.22) | 82 (461) | 689 | 90.50 | 98.24 (92.97) | 128 (668) | 1364 |
|  | T&T | | | | CPC | | | |
| SIFT (python) | 60.30 | 66.21 (46.66) | 84 (548) | 4975 | 25.20 | 59.17 (38.59) | 46 (270) | 4650 |
| Key.Net[2]+HardNet[3] | 69.20 | 76.48 (72.97) | 80 (666) | 5000 | 32.50 | 65.43 (65.65) | 58 (286) | 5000 |
| ASLFeat [1] | 82.30 | 83.99 (79.26) | 116 (668) | 4999 | **49.30** | 88.66 (84.83) | 74 (356) | 4753 |
| SobelNet | **82.60** | 83.95 (73.90) | 85 (856) | 4943 | 41.80 | 81.00 (72.67) | 47 (365) | 4469 |

*B. FM-Bench*

While HPatches benchmark mainly concentrates on homography transformation and matching task, FM-Bench proposes a geometry estimation pipeline which includes pruning bad correspondences by Lowe's ratio test [14] and outlier rejection by robust estimators such as RANSAC [15]. This pipeline finally generates the estimated fundamental matrices, which can effectively represent the ability of the keypoints and their descriptors to recover the pose transformation matrix of the camera, between the corresponding images.

The benchmark datasets include: 1) The TUM dataset [18], which provides videos of indoor scenes, where the texture is often weak and images are sometimes blurred due to the fast camera movement. 2) The KITTI odometry dataset [19], which consists of consecutive frames in a driving scenario, where the geometry between images is dominated by the forward motion. 3) The Tanks and Temples (T&T) dataset [20], which provides many scans of scenes or objects for image-based reconstruction, and hence offers wide-baseline pairs for evaluation. 4) The Community Photo Collection (CPC) dataset [21], which provides unstructured images of well-known landmarks across the world collected from Flickr. 1000 pairs in each dataset are randomly chosen for testing.

Notice that the TUM images used in this benchmark are different from the training and valuation dataset of SobelNet.

The most important metric in FM-Bench is %Recall. Given the fundamental matrix estimates, they are classified as accurate or not by thresholding the Normalized SGD error (0.05 by default), and use the %Recall, the ratio of accurate estimates to all estimates, for evaluation. The benchmark also presents the inlier rate (%inlier), i.e., the ratio of inliers to all matches, and the inlier rate before outlier rejection (%inlier-m). Besides, correspondence numbers (#corrs), which is less important since the impact of match numbers to high-level applications such as SfM [16] are arguable [17] are presented while #corrs-m means the match numbers before the estimation pipeline. In this section, we limit the maximum number of keypoints of all methods to 5K and set NMS size 3 for machine learning methods the same as [1]. The experiment results are shown in Table II. Our method works very well in the TUM, KITTI, and T&T datasets in which images are transformed continuously, but not so well in the CPC dataset in which images are extremely transformed.

*C. ETH benchmark*

We exploit the ETH benchmark to evaluate the performance of our method in terms of 3D reconstruction tasks. We evaluate our method on three small-scale benchmark datasets, the well-known MVS benchmark [23], and the South Building dataset [24]. The number of registered images and sparse points quantifies the completeness of the reconstruction. The number of observations per image, i.e., the

Table III    Evaluation on ETH Benchmark

| Dataset | method | #Reg. Images | #Sparse Points | #Observations | Track Length | Obs. Per Image | Reproj. Error[px] | #Dense Point | # Inlier Pairs | #Inlier Matches |
|---|---|---|---|---|---|---|---|---|---|---|
| Fountain (11) | RootSIFT[22] | 11 | 14722 | 70631 | 4.79765 | 6421.00 | **0.392893** | 292609 | **55** | 127734 |
| | Key.Net [2] + HardNet [3] | 11 | 20792 | 109142 | **5.24923** | 9922.00 | 0.633731 | **306151** | 53 | 193979 |
| | ASLFeat [1] | 11 | 21933 | 110183 | 5.023617 | 10016.64 | 1.039979 | 540 | 54 | 237034 |
| | SobelNet | 11 | **25120** | **130281** | 5.186346 | **11843.73** | 0.652734 | 305583 | **55** | 240876 |
| Herzjesu (8) | RootSIFT[22] | 8 | 7502 | 31670 | 4.22154 | 3958.75 | **0.431632** | 241347 | 28 | 48965 |
| | Key.Net [2] + HardNet [3] | 8 | 13412 | 57562 | 4.291828 | 7195.25 | 0.61847 | 239164 | 28 | 86419 |
| | ASLFeat [1] | 8 | 9654 | 38592 | 3.997514 | 4824.0 | 1.010794 | 821 | 28 | 77567 |
| | SobelNet | 8 | **16557** | **74240** | **4.483904** | **9280.0** | 0.663757 | 241339 | 28 | **120677** |
| South-Building (128) | RootSIFT[22] | 128 | 108124 | 953975 | 6.04838 | 5109.18 | **0.545747** | 2141964 | 3822 | 2036024 |
| | Key.Net [2] + HardNet [3] | 128 | 138736 | **1278271** | 9.213694 | 9986,49 | 0.679462 | **2148250** | 5089 | 5539715 |
| | ASLFeat [1] | 127 | **180802** | 1109031 | 6.133953 | 8732.53 | 1.187762 | 7171 | 3005 | **6443388** |
| | SobelNet | 128 | 163474 | 1257644 | 7.693236 | 9825.34 | 0.775618 | 2125289 | **5206** | 4815976 |

Table IV    Efficiency comparison

| method | Detection | Descriptor | Detection score | Post process |
|---|---|---|---|---|
| Key.Net | $3.659*10^9$ | None | output | False |
| ASLFeat | $19.906*10^9$ | | peakiness | True |
| SobelNet | $1.889*10^9$ | $4.819*10^9$ | norm_output | False |

*As for ASLFeat, the number does not include the additional computation required for DCN [28].

number of verified image projections of sparse points, and the track length, i.e., the number of verified image observations per sparse point is crucial for accurate calibration of the cameras and reliable triangulation. The overall reprojection error indicates the accuracy of the reconstruction. The dense point number is a single measure of the overall completeness of the reconstruction and the accuracy of the SFM results. In this section, we limit the maximum number of keypoints of all methods to 20K and set NMS size 3 for machine learning methods the same as [1]. The experiment results are shown in Table III. RootSIFT shows the best performance on reprojection error while machine learning methods always have a longer track length.

### D. Efficiency

The inference time of an image of 640×480 is **7.59ms** and **1.09ms** for SobelNet and DesNet respectively on RTX 2070 SUPER. The efficiency comparison of multiplication operation numbers among our method, Key.Net, and ASLFeat is shown in Table IV. The computation of our method is much less than the other two. Besides, ASLFeat needs to compute peakiness score and post-processes after convolutional layers which could lead to a longer inference time.

### VI. CONCLUSIONS

In this paper, we propose SobelNet trained by Gauss loss and DesNet trained by circle loss to detect keypoints and compute corresponding descriptors. Compared with previous machine learning [1, 2] and traditional [7] **SOTA** methods, SobelNet has a better or comparable performance on several benchmarks. Our method can work at high frequency on a common device due to our lightweight network architecture. However, we meet a similar problem as SuperPoint [4]. As shown in Fig. 5, SobelNet performs badly with default params owing to an extreme rotation angle. This problem could be solved with more careful data augmentation. For example, we rotate the images by [-90, 90] degrees randomly during the training process of DesNet. As a result, DesNet performs well on image pairs with extreme rotations but worse on the above benchmarks in which images do not rotate so much. Overall, if the camera moves continuously and slightly, it's OK to use our method (in chapter IV) to train DesNet, but if the camera moves at high speed or rotates too fast, we need to train DesNet more carefully.

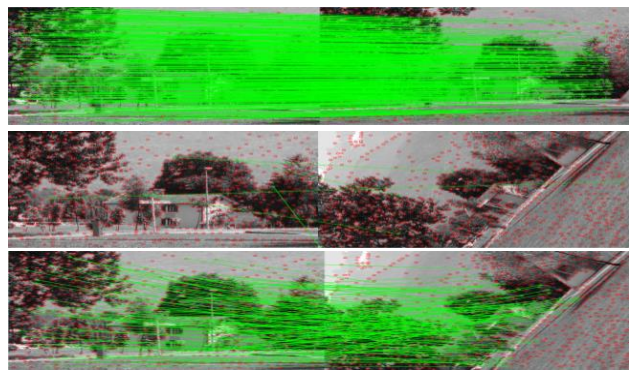

Fig. 5. The upper pair of images shows a good match with small transformation. The middle pair shows a bad match due to the extreme rotation. The bottom pair shows a relatively good match using parameters which are trained deliberately for the image rotation between [-90, 90] degrees.

## REFERENCES

[1] Luo, Zixin, et al. "Aslfeat: Learning local features of accurate shape and localization." *Proceedings of the IEEE/CVF Conference on Computer Vision and Pattern Recognition*. 2020.

[2] Barroso-Laguna, Axel, et al. "Key. net: Keypoint detection by handcrafted and learned cnn filters." *Proceedings of the IEEE International Conference on Computer Vision*. 2019.

[3] Mishchuk, Anastasiia, et al. "Working hard to know your neighbor's margins: Local descriptor learning loss." *Advances in Neural Information Processing Systems*. 2017.

[4] DeTone, Daniel, Tomasz Malisiewicz, and Andrew Rabinovich. "Superpoint: Self-supervised interest point detection and description." *Proceedings of the IEEE Conference on Computer Vision and Pattern Recognition Workshops*. 2018.

[5] Christiansen, Peter Hviid, et al. "Unsuperpoint: End-to-end unsupervised interest point detector and descriptor." *arXiv preprint arXiv:1907.04011* (2019).

[6] Rublee, Ethan, et al. "ORB: An efficient alternative to SIFT or SURF." *2011 International conference on computer vision*. Ieee, 2011.

[7] Lowe, David G. "Distinctive image features from scale-invariant keypoints." *International journal of computer vision* 60.2 (2004): 91-110.

[8] Revaud, Jerome, et al. "R2d2: Repeatable and reliable detector and descriptor." *arXiv preprint arXiv:1906.06195* (2019).

[9] Ono, Yuki, et al. "LF-Net: learning local features from images." *Advances in neural information processing systems*. 2018.

[10] Sun, Yifan, et al. "Circle loss: A unified perspective of pair similarity optimization." *Proceedings of the IEEE/CVF Conference on Computer Vision and Pattern Recognition*. 2020.

[11] Schonberger, Johannes L., et al. "Comparative evaluation of hand-crafted and learned local features." *Proceedings of the IEEE Conference on Computer Vision and Pattern Recognition*. 2017.

[12] Bian, Jia-Wang, et al. "An evaluation of feature matchers for fundamental matrix estimation." *arXiv preprint arXiv:1908.09474* (2019).

[13] Balntas, Vassileios, et al. "HPatches: A benchmark and evaluation of handcrafted and learned local descriptors." *Proceedings of the IEEE Conference on Computer Vision and Pattern Recognition*. 2017.

[14] David G Lowe. Distinctive image features from scale-invariant keypoints. *International Journal on Computer Vision (IJCV)*, 60(2):91–110, 2004.

[15] Martin A Fischler and Robert C Bolles. Random sample consensus: a paradigm for model fitting with applications to image analysis and automated cartography. *Communications of the ACM*, 24(6):381–395, 1981.

[16] Johannes L Schönberger and Jan-Michael Frahm. Structure-from-motion revisited. In *IEEE Conference on Computer Vision and Pattern Recognition (CVPR)*, pages 4104–4113, 2016.

[17] Johannes L Schönberger, Hans Hardmeier, Torsten Sattler, and Marc Pollefeys. Comparative evaluation of hand-crafted and learned local features. In *IEEE Conference on Computer Vision and Pattern Recognition (CVPR)*, pages 6959–6968. IEEE, 2017.

[18] J. Sturm, N. Engelhard, F. Endres, W. Burgard, and D. Cremers. A benchmark for the evaluation of RGB-D SLAM systems. In *IEEE International Conference on Intelligent Robots and Systems (IROS)*, Oct. 2012.

[19] Andreas Geiger, Philip Lenz, and Raquel Urtasun. Are we ready for autonomous driving? the KITTI vision benchmark suite. In *IEEE Conference on Computer Vision and Pattern Recognition (CVPR)*, pages 3354–3361. IEEE, 2012.

[20] Arno Knapitsch, Jaesik Park, Qian-Yi Zhou, and Vladlen Koltun. Tanks and Temples: Benchmarking large-scale scene reconstruction. *ACM Transactions on Graphics (TOG)*, 36(4):78, 2017.

[21] Kyle Wilson and Noah Snavely. Robust global translations with 1DSFM. In European Conference on Computer Vision (ECCV), pages 61–75. Springer, 2014.

[22] Arandjelović, Relja, and Andrew Zisserman. "Three things everyone should know to improve object retrieval." *2012 IEEE Conference on Computer Vision and Pattern Recognition*. IEEE, 2012.

[23] Strecha, Christoph, et al. "On benchmarking camera calibration and multi-view stereo for high resolution imagery." *2008 IEEE Conference on Computer Vision and Pattern Recognition*. Ieee, 2008.

[24] Hane, Christian, et al. "Joint 3D scene reconstruction and class segmentation." *Proceedings of the IEEE Conference on Computer Vision and Pattern Recognition*. 2013.

[25] Mur-Artal, Raul, and Juan D. Tardós. "Orb-slam2: An open-source slam system for monocular, stereo, and rgb-d cameras." *IEEE Transactions on Robotics* 33.5 (2017): 1255-1262.

[26] Carmigniani, Julie, et al. "Augmented reality technologies, systems and applications." *Multimedia tools and applications* 51.1 (2011): 341-377.

[27] Tian, Yurun, et al. "D2D: Keypoint Extraction with Describe to Detect Approach." *arXiv preprint arXiv:2005.13605* (2020).

[28] Dai, Jifeng, et al. "Deformable convolutional networks." *Proceedings of the IEEE international conference on computer vision*. 2017.

[29] Lin, Tsung-Yi, et al. "Microsoft coco: Common objects in context." *European conference on computer vision*. Springer, Cham, 2014.